\documentclass[fleqn,10pt]{wlscirep}
\usepackage[utf8]{inputenc}
\usepackage[T1]{fontenc}
\usepackage[detect-all]{siunitx}
\usepackage{subcaption}
\usepackage{xspace}
\newcommand{\acinoset}{AcinoSet\xspace}
\newcommand{\dlc}{DeepLabCut\xspace}
\newcommand{\fte}{FTE\xspace}
\newcommand{\kfte}{K-FTE\xspace}

\newcommand{\vect}[1]{\mathbf{#1}}
\newcommand{\dvect}[1]{\mathbf{\dot{#1}}}
\newcommand{\ddvect}[1]{\mathbf{\ddot{#1}}}

\newcommand{\ie}{i.e.\ }
\newcommand{\eg}{e.g.\ }

\title{Wild Motion Unleashed: Markerless 3D Kinematics and Force Estimation in Cheetahs}
\author[1,*]{Zico da Silva}
\author[1]{Stacy Shield}
\author[2]{Penny E. Hudson}
\author[3]{Alan M. Wilson}
\author[1]{Fred Nicolls}
\author[1]{Amir Patel}
\affil[1]{University of Cape Town, Department of Electrical Engineering, Cape Town, 7700, South Africa}
\affil[2]{University of Chichester, Institute of Sport Nursing and Allied Health, Chichester, PO19 6PE, United Kingdom}
\affil[3]{The Royal Veterinary College, Structure and Motion Laboratory, London, NW1 0TU, United Kingdom}

\affil[*]{zicods7@gmail.com}

\begin{abstract}
The complex dynamics of animal manoeuvrability in the wild is extremely challenging to study. The cheetah (\textit{Acinonyx jubatus}) is a perfect example: despite great interest in its unmatched speed and manoeuvrability, obtaining complete whole-body motion data from these animals remains an unsolved problem. This is especially difficult in wild cheetahs, where it is essential that the methods used are remote and do not constrain the animal's motion. In this work, we use data obtained from cheetahs in the wild to present a trajectory optimisation approach for estimating the 3D kinematics and joint torques of subjects remotely. We call this approach kinetic full trajectory estimation (K-FTE) . We validate the method on a dataset comprising synchronised video and force plate data. We are able to reconstruct the 3D kinematics with an average reprojection error of \qty{17.69}{pixels} (\qty{62.94}{\percent} PCK using the nose-to-eye(s) length segment as a threshold), while the estimates produce an average root-mean-square error of \qty{171.3}{\newton} ($\approx \qty{17.16}{\percent}$ of peak force during stride) for the estimated ground reaction force when compared against the force plate data. While the joint torques cannot be directly validated against ground truth data, as no such data is available for cheetahs, the estimated torques agree with previous studies of quadrupeds in controlled settings. These results will enable deeper insight into the study of animal locomotion in a more natural environment for both biologists and roboticists.
\end{abstract}
\begin{document}

\flushbottom
\maketitle
%
%
\thispagestyle{empty}
\section*{Introduction}

High-speed manoeuvrability is the "final frontier" in the study of legged locomotion. While constant-speed gait has been studied extensively, the dynamics and control of these transient movements are still sparsely investigated. This could be attributed to the complex dynamics manoeuvring entails, which require whole-body motion and force data to quantify.
This data can be gathered in a lab setting~\cite{robertson2013research} but does not reflect the natural locomotor conditions or animal motivations experienced in the field. Rapid manoeuvres are important to understand, however. From a biomechanics perspective, they push animals to perform at their mechanical limits, revealing what they are capable of in the most extreme circumstances. They are also interesting to the robotics field, as understanding this category of motion will be crucial for the development of more agile robotic systems that better match the capabilities of animals ~\cite{daley2016non}.

The cheetah (\textit{Acinonyx jubatus}) is the perfect model for studying quadruped dynamics as it is not only the fastest terrestrial animal, but also one of the most manoeuvrable. In fact, a study which used GPS-IMU collars to investigate the behaviour of wild cheetahs revealed that it is the ability to rapidly accelerate that is most critical to their hunting success~\cite{wilson2013locomotion}.
Tracking collars are, however, only able to treat the animal as one lumped rigid body and are thus unable to provide information about leg, spine or tail kinematics or joint loading.

By contrast, vision-based pose estimation methods provide a means for non-invasive full-body kinematic and kinetic estimation (``dynamic'' estimation).
In human research, this technique has been adopted to obtain 3D pose estimates from a single camera, \ie monocular 3D pose estimation~\cite{mehta2017monocular, martinez2017simple}, focusing on full-body kinematics.
While data-driven models or purely kinematic formulations are widely applied in research of human locomotion, there have been efforts to incorporate a more complete physics-based model~\cite{li2022estimating, shimada2020physcap}, which often is formulated as a non-linear program (NLP) to solve for both the kinematic and kinetic parameters.

These models provide a strong prior on the motion, making it a popular choice for potentially ambiguous pose detection from a single camera setup. The internal and external torques and forces are a natural byproduct of modelling the kinetics together with the kinematics. Some researchers have used this to explicitly analyse joint torques produced by humans while performing different activities~\cite{riemer2008improving, zell2017joint}.
In conjunction, optimal-control-based approaches have been used to fit a human model to corresponding kinematic data to obtain joint torques and contact forces~\cite{felis2015optimal, schemschat2016joint}.

For all the work done on humans, there is little to show for markerless dynamic motion estimation of animals in the wild. That said, animal 3D pose estimation (without explicit modelling of physics), for both multi-view camera and monocular systems, have been well established in recent years. The multi-view camera techniques often resort to triangulation-based methods~\cite{nathUsingDeepLabCut3D2019, karashchukAniposeToolkitRobust2021, joska2021acinoset}, while the monocular systems have used the skinned multi-animal linear model (SMAL)~\cite{zuffi20173d, biggs2018creatures} and pose ``lifting''~\cite{gosztolai2021liftpose3d} to disambiguate 3D pose estimates from a single view.
These methods have been successful at 3D pose estimation of animals in the wild, but they lack the joint torques and ground reaction forces (GRFs) that are important for biomechanic analysis and robotic design.

Researchers have estimated animal torques and GRFs using invasive methods in controlled experiments.
In one study, researchers placed reflective markers on racing greyhounds and, together with a six-camera setup and force plates, were able to determine joint torque and power profiles of the hind limbs~\cite{williams2009exploring}. Similar work was done on the sit-to-stand dynamics of greyhounds~\cite{ellis2018limb}. In the same line  of work, a study was done on the forelimb muscle activity of horses~\cite{harrison2012forelimb}.

While there are examples of dynamic motion estimation being applied to animals, work has mostly been limited to a laboratory setting, or involved invasive markers (or trackers) being placed on the subjects in the wild. In contrast, the proposed method performs markerless full-body kinetic and kinematic estimation of animals (in this case, cheetahs) with a multi-camera setup. To the best of our knowledge, this is a novel approach to the dynamic estimation of wildlife locomotion in the wild.

In our previous work, \acinoset~\cite{joska2021acinoset, muramatsu2022improving}, we were able to reconstruct the 3D kinematics of free-running cheetahs from multiple cameras. Here, we extend this work to include kinetic modelling for the cheetah using a complete physics-based model to describe the motion.
In particular, we are interested in the joint torques during a gallop.
To achieve this, we created a new dataset (denoted ``kinetic dataset'' in this work) that contains synchronised video and force plate data.
As with \acinoset, a trajectory optimisation problem (henceforth referred to as the kinetic full trajectory estimation (\kfte) method\footnote{This differs from the purely kinematic full trajectory estimation (\fte) method developed for \acinoset.}) is formulated and solved for the dynamic motion estimate, and then evaluated against the kinetic dataset. Once the \kfte method has been validated with the kinetic dataset, we can use this method to perform motion and torque estimation on a test set of \acinoset. This provides valuable information about the joint loading of the limbs of the cheetah during locomotion.

Even though this work focuses on the cheetah, the proposed method can easily be generalised to work with any wild animal. This will enable biologists to understand animal locomotion in much more detail and can provide data for the design of new robotic controllers.

\section*{Results}
We first present the quantitative and qualitative results of the motion and torque estimation on both datasets, and then we motivate its use using the kinetic dataset to compare estimated GRFs with the ground truth measurements from the force plates.
\subsection*{Motion and Torque Estimation}
The 3D pose estimation results are shown in Table~\ref{tab:torques}.
The mean position error (MPE) provides a relative measure of how well the reconstruction matches the baseline \fte result using our previous kinematic methods~\cite{joska2021acinoset, muramatsu2022improving}.
The MPE is minimal, suggesting that the \kfte method produces similar 3D kinematics to the baseline \fte.

Note that we calculated an average reprojection error of \qty{17.69}{pixels} (\qty{62.94}{\percent} PCK using the nose-to-eye(s) length threshold) using a subset of \qty{540}{frames} of 2D hand-labelled data as ground truth from two different trials, T4 and T5.
This result compliments the MPE shown in Table~\ref{tab:torques}, while also providing an absolute measure of the accuracy of the 3D reconstruction.
\begin{table}[htbp]
\begin{center}
\caption{Error analysis of the resultant 3D reconstruction on \acinoset that includes a physics-based model for dynamic data estimation.}\label{tab:torques}%
\begin{tabular}{@{}lcc@{}}
\toprule
Test & MPE (\unit{\mm})\\
\midrule
T1 & $17.6$\\
T2 & $34.0$\\
T3 & $19.2$\\
T4 & $20.7$\\
T5 & $23.5$\\
\bottomrule
\end{tabular}
\end{center}
\end{table}

\begin{figure}[htpb]
\centering
\includegraphics[scale=0.25]{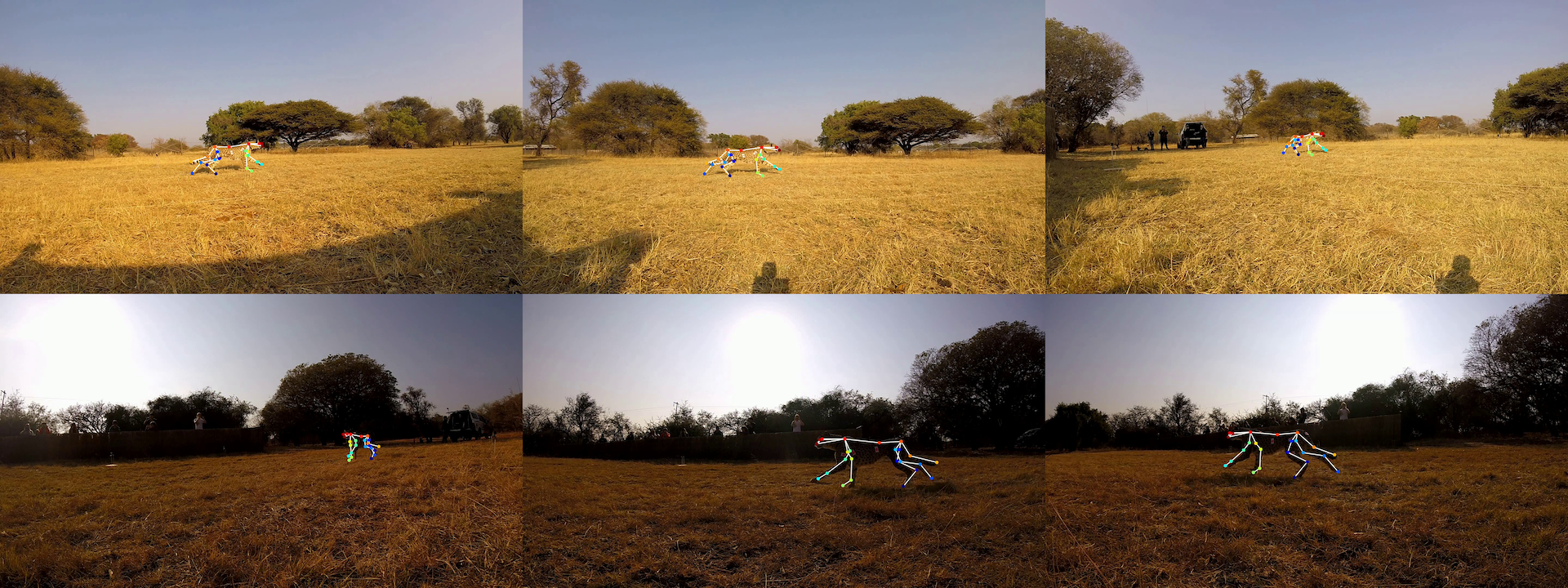}
\caption{An visual example of a single pose viewed from all \qty{6} camera angles for T1 of \acinoset. This illustrates good 3D pose estimates of the cheetah.}\label{fig:qualitative}
\end{figure}

Figure~\ref{fig:qualitative} shows a single pose viewed from all \qty{6} camera angles.
Visually, the pose estimate is reasonably accurate given the estimate is well-matched to the cheetah's skeleton in each view.

\begin{figure*}[htbp]
\centering
\begin{subfigure}{0.5\textwidth}
  \centering
  \includegraphics[width=\textwidth]{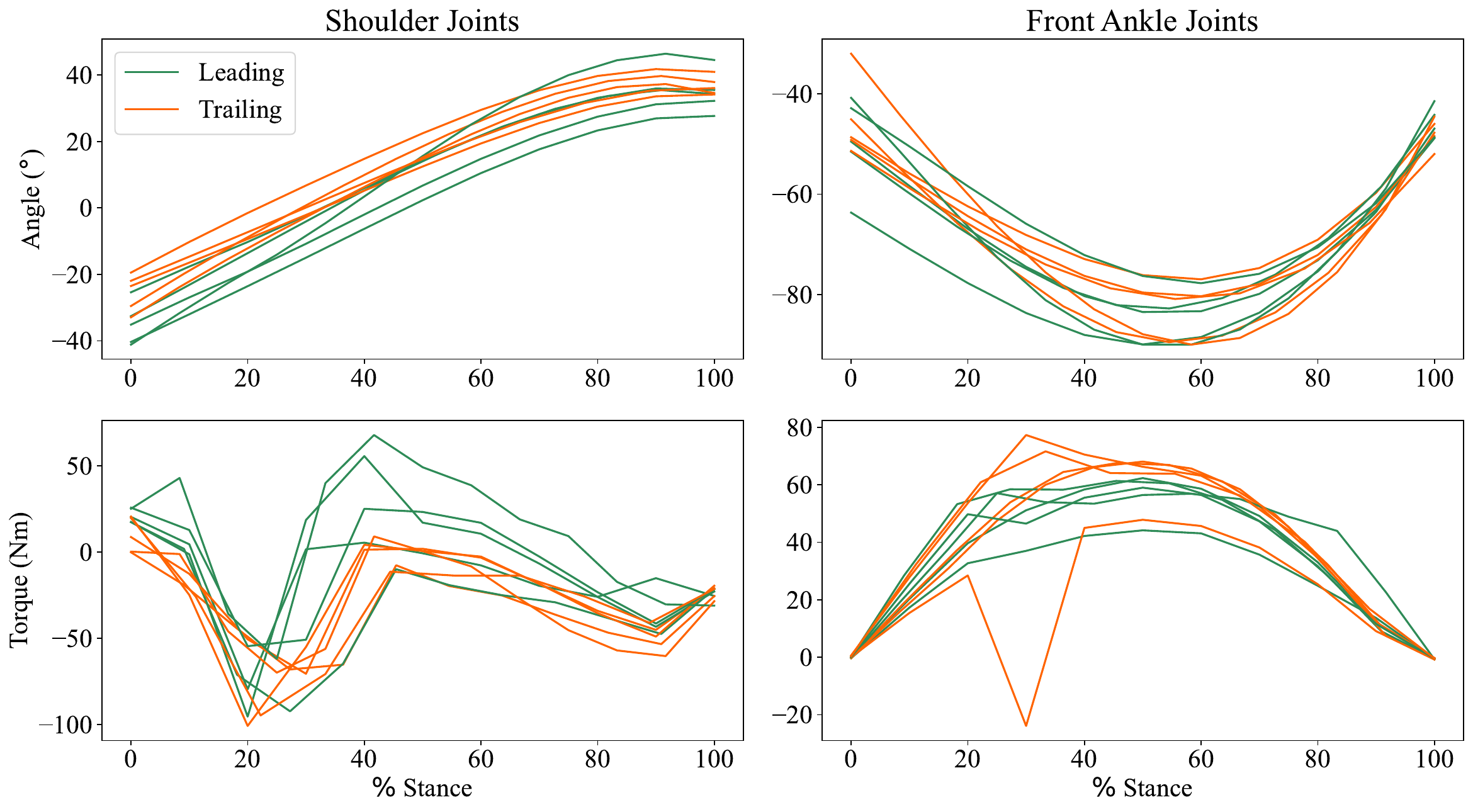}
  \caption{}
  \label{fig:kinetic_forelimb}
\end{subfigure}%
\begin{subfigure}{0.5\textwidth}
    \centering
    \includegraphics[width=\textwidth]{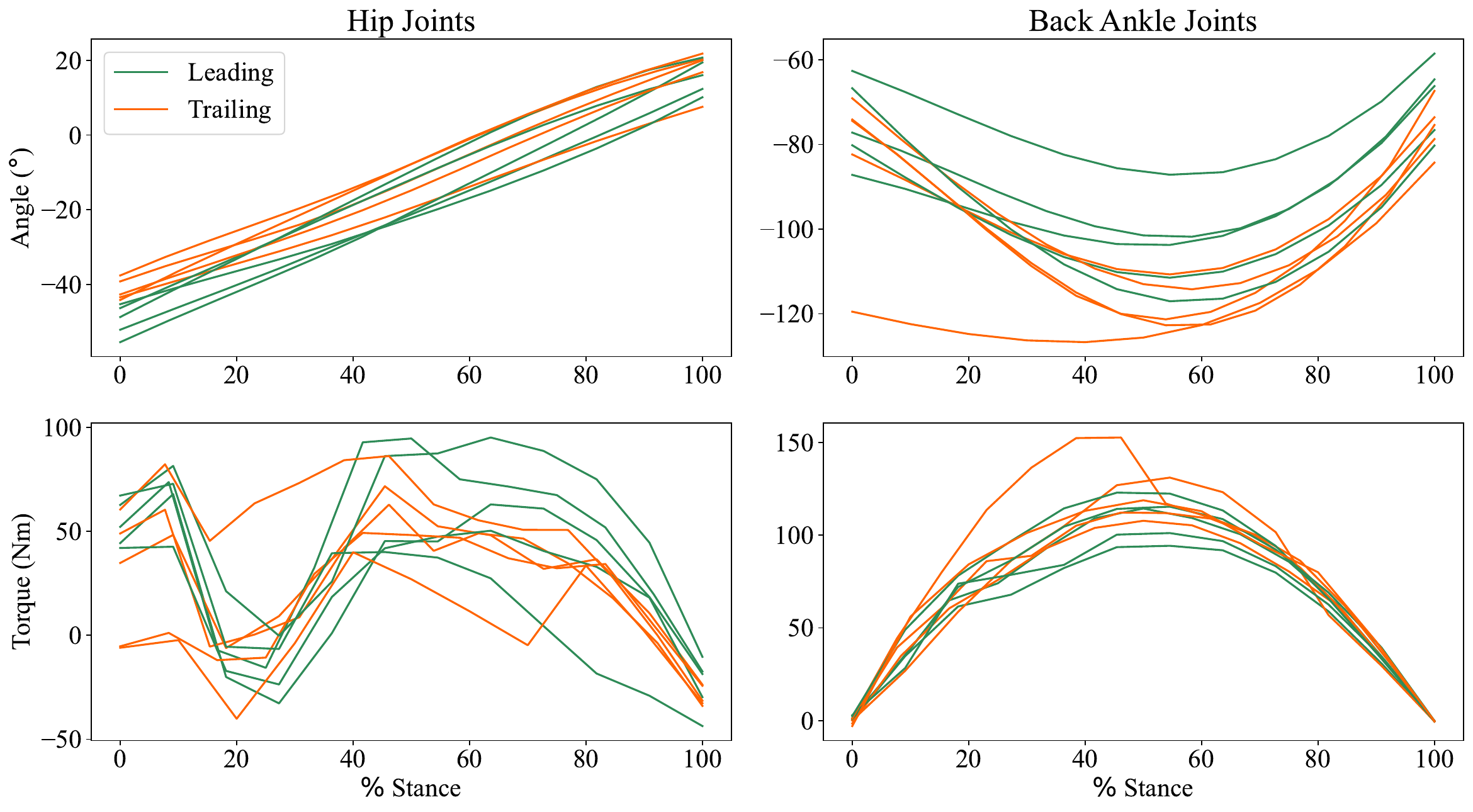}
    \caption{}
    \label{fig:kinetic_hindlimb}
\end{subfigure}
\caption{Joint angle and torque estimates of the forelimb (\subref{fig:kinetic_forelimb}) and hindlimb (\subref{fig:kinetic_hindlimb}) hip and hock joints during stance phase of gait for the kinetic subset of the evaluation dataset. The top rows are the joint angles and bottom rows the joint torques. The joint angles and torques are consistent during the stance phase.}
\label{fig:kinetic_torque_analysis}
\end{figure*}

\begin{figure*}[htbp]
\centering
\begin{subfigure}{0.5\textwidth}
  \centering
  \includegraphics[width=\textwidth]{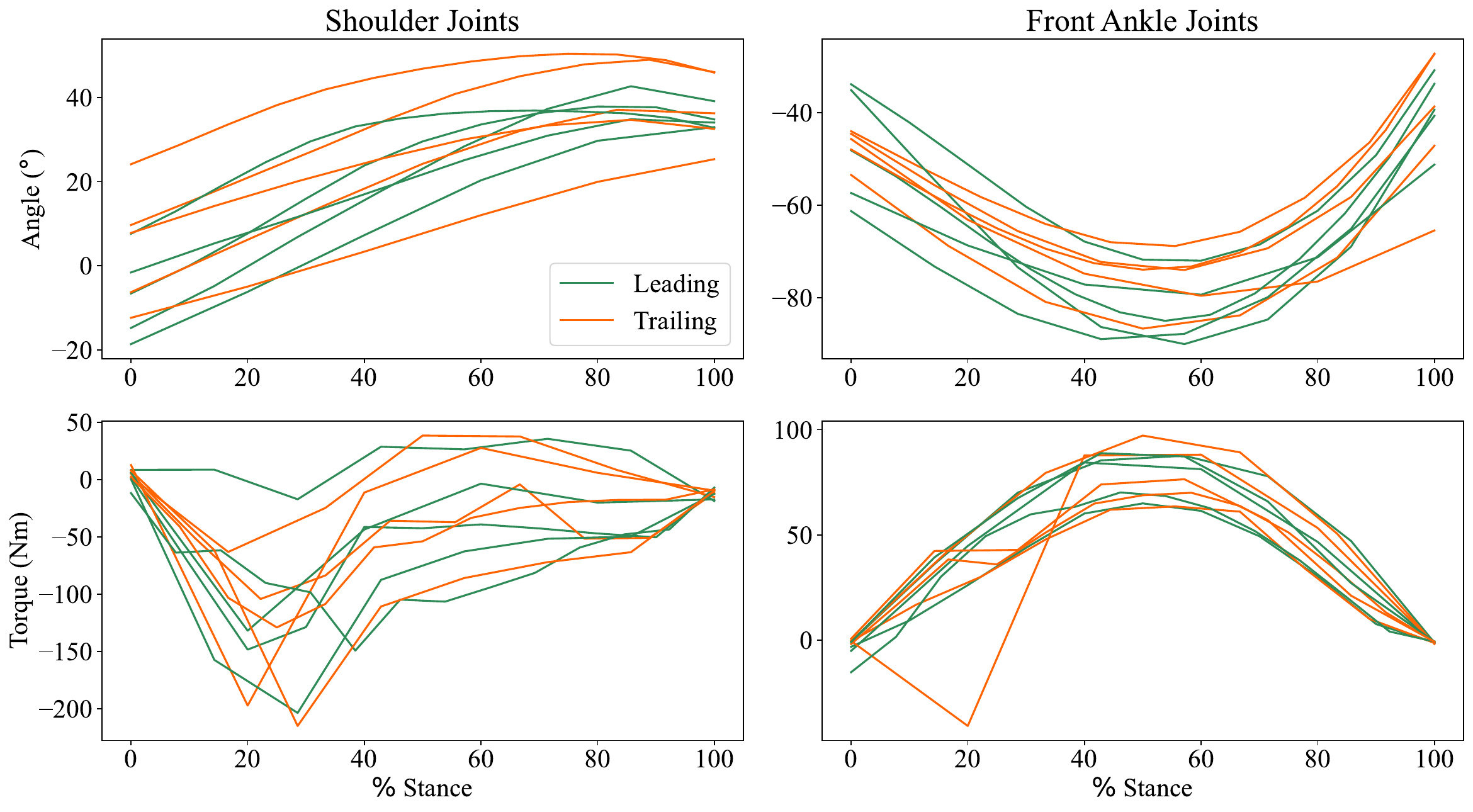}
  \caption{}
  \label{fig:acinoset_forelimb}
\end{subfigure}%
\begin{subfigure}{0.5\textwidth}
    \centering
    \includegraphics[width=\textwidth]{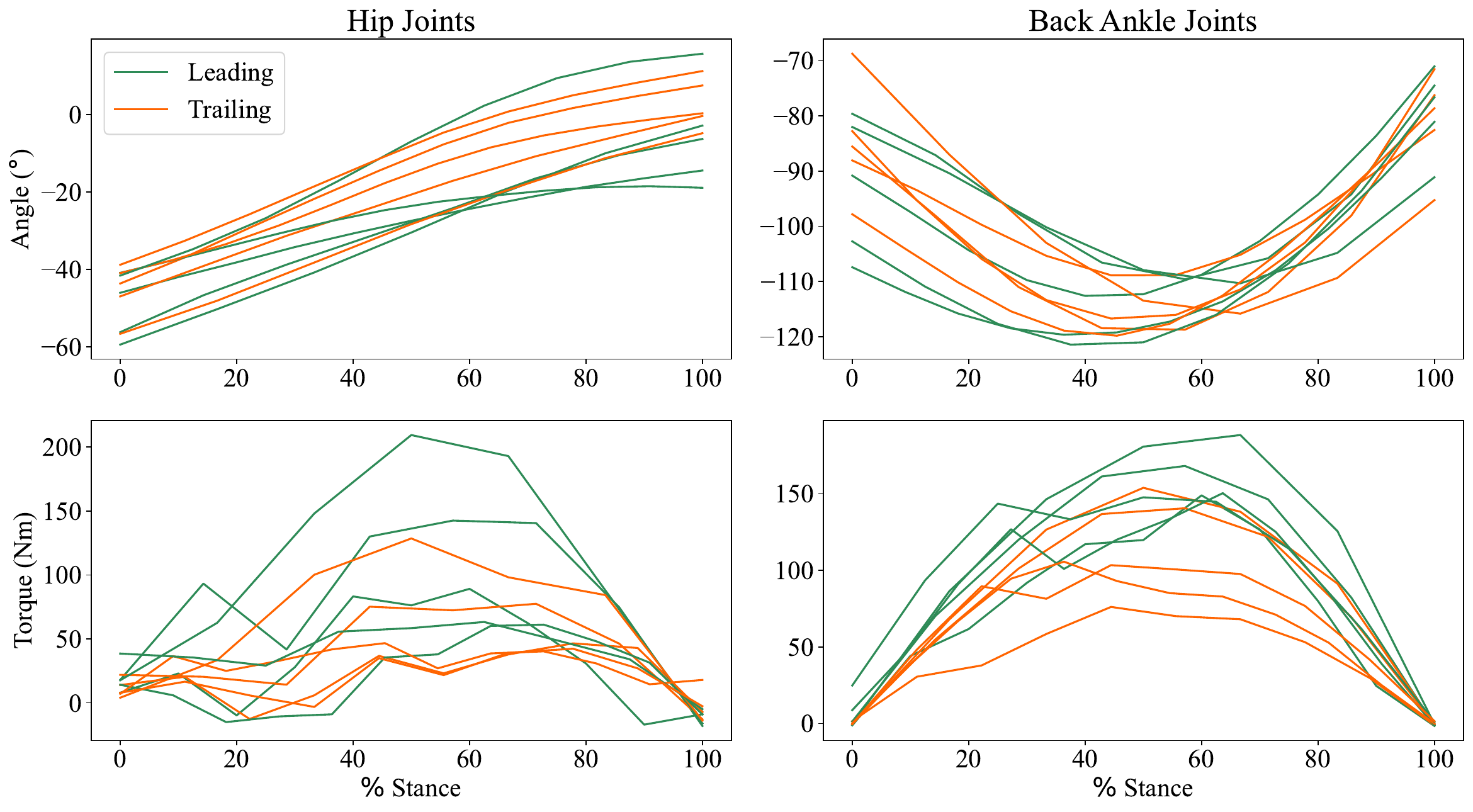}
    \caption{}
    \label{fig:acinoset_hindlimb}
\end{subfigure}
\caption{Joint angle and torque estimates of the forelimb (\subref{fig:acinoset_forelimb}) and hindlimb (\subref{fig:acinoset_hindlimb}) hip and hock joints during stance phase of gait for \acinoset subset of the evaluation dataset. The top rows are the joint angles and bottom rows the joint torques. The joint angles and torques are consistent during the stance phase.}
\label{fig:acinoset_torque_analysis}
\end{figure*}

The estimated joint angles and torques for the fore and hind limbs, while in contact with the ground (stance phase of the gait), are shown in Figures~\ref{fig:kinetic_torque_analysis} and~\ref{fig:acinoset_torque_analysis} for the kinetic dataset and \acinoset respectively.
The torque estimates produced for the kinetic dataset serve as a reference for what is expected given that we know the corresponding GRF.
There is a good correlation between the torque estimates during the stance for both datasets.

Based on the kinetic dataset (see Figure~\ref{fig:kinetic_torque_analysis}), both the fore and hind limbs exhibit similar torque trajectories. For the front and back ankle joints, the torque profile during the stance follows a half sine wave, where the peak torque on average is around \qty{50}{\percent}. The hindlimb produces a slightly higher peak torque on average than the forelimb.

The hip torques do not conform to as consistent a profile as the ankle torques do, but they are consistently bounded between \qty{100}{\newton\metre} and \qty{-50}{\newton\metre}. The trailing shoulder joint torques produce peaks at around \qty{20}{\percent} of the stance, and quickly returns to zero thereafter. A similar pattern is found in the trailing shoulder joint, however the torque peaks on average around two points in the stance: a negative peak at \qty{20}{\percent} and a positive peak at \qty{40}{\percent}.

There is consistency in the joint angles for the duration of the stance. The shoulder and hip joints increase linearly from start to finish, while the front and back ankle joints display a parabola-like change during the stance.

\subsection*{Method Validation}
It must be stressed that the force and torque values obtained using this method are \textit{estimates} at best -- not remote measurements. If multiple feet are on the ground, the inverse dynamics problem does not have a unique solution, so the results are merely one possible combination of forces out of many that could produce the observed motion in the 3D model. Still, these estimates should give a reliable indication of the magnitudes of the forces and torques involved, even if the precise profiles cannot be confirmed.

Without ground truth joint torques, we set up an alternative evaluation procedure to determine whether the observed joint torques are plausible. We make use of the measurements from the force plate as a proxy for a ``ground truth'' to validate motion and torque estimates. Two \kfte methods for estimating the ground reaction forces were compared: sinusoidal GRF and freeform GRF. The sinusoidal approach assumes a ``sinusoidal'' GRF profile, while the freeform approach does not make any assumptions about the GRF profile.
The idea of using a sinusoidal profile for the GRF was taken from a previous study that examined the GRFs on horses during steady-state galloping~\cite{witte2004determination}.
In addition, the force plate obtained GRF profiles found in the kinetic dataset exhibited the familiar sinusoidal shape, and therefore, it made sense to develop a method that uses this inherent structure.

\begin{figure*}[htbp]
\centering
\begin{subfigure}{0.5\textwidth}
  \centering
  \includegraphics[width=\textwidth]{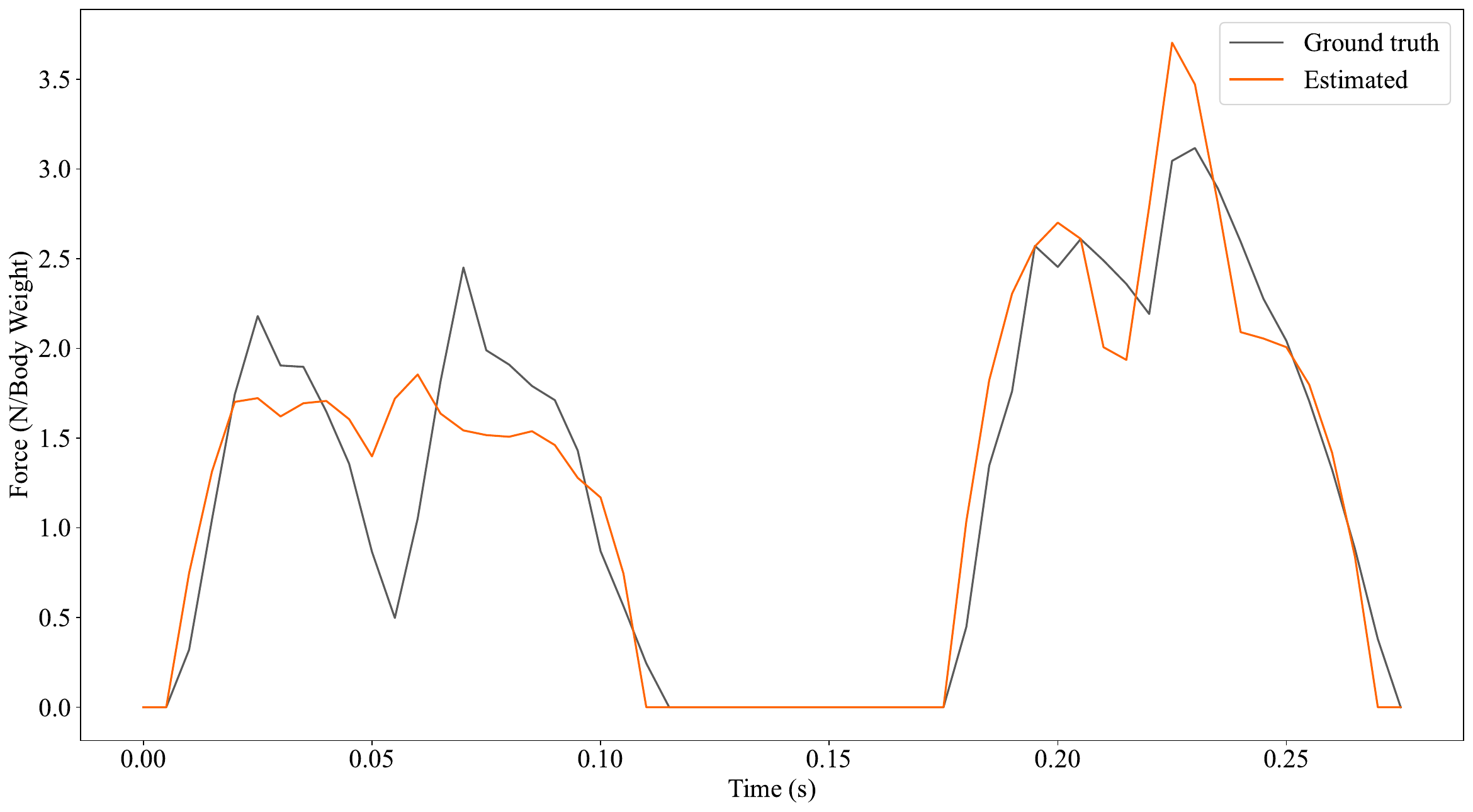}
  \caption{}
  \label{fig:grf_1}
\end{subfigure}%
\begin{subfigure}{0.5\textwidth}
    \centering
    \includegraphics[width=\textwidth]{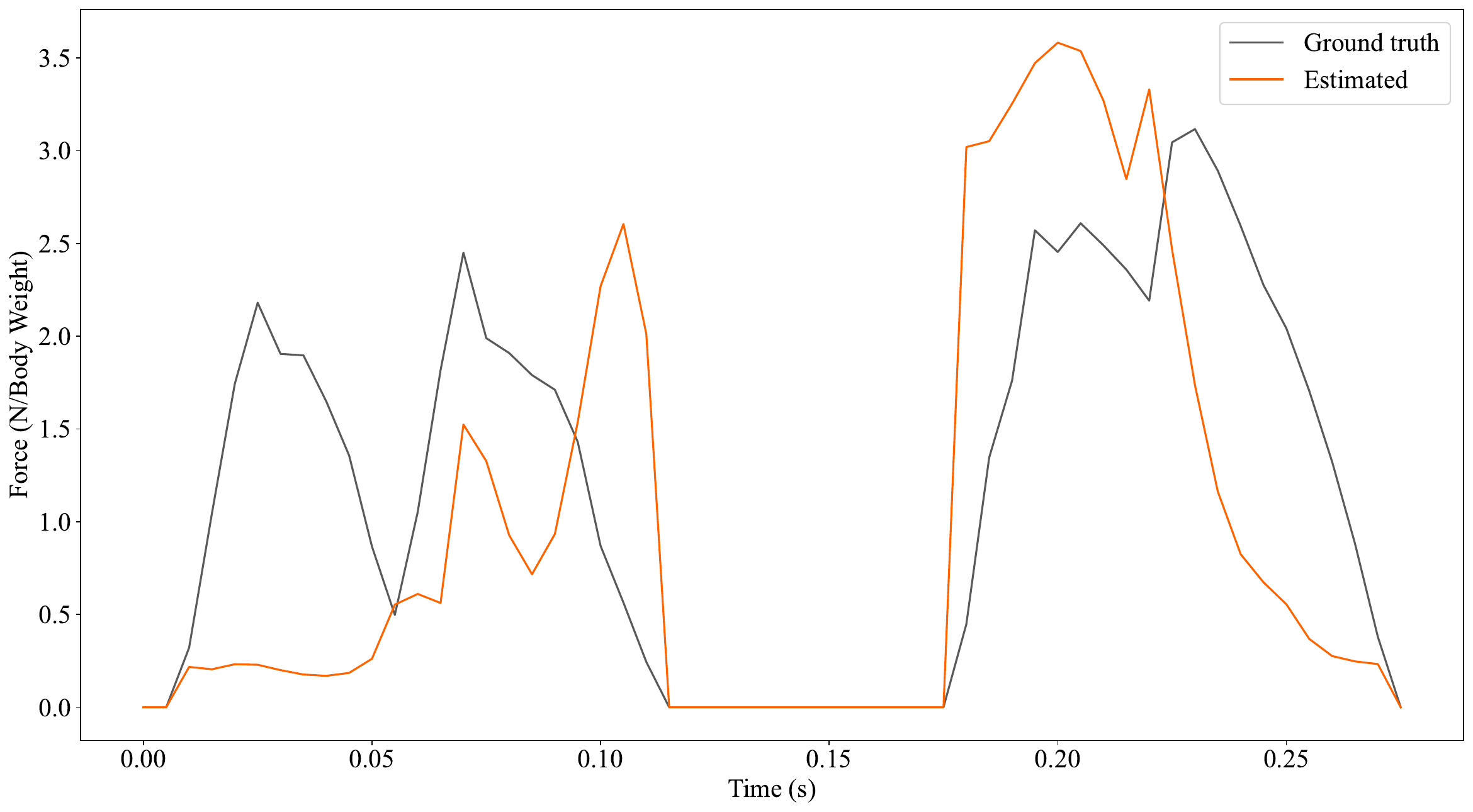}
    \caption{}
    \label{fig:grf_2}
\end{subfigure}
\caption{A visual example comparing the resultant GRF magnitudes produced by both methods: (\subref{fig:grf_1}) sinusoidal GRF, and (\subref{fig:grf_2}) freeform GRF. This is a result from T4 of the kinetic dataset. The force has been normalised by the body weight of the subject.}
\label{fig:grf_estimation}
\end{figure*}

Table~\ref{tab:grf} provides the GRF estimation error for both methods, and Figure~\ref{fig:grf_estimation} provides a visual example of the GRFs.
\begin{table}[htbp]
\begin{center}
\caption{GRF estimation comparison between the estimated and the force plate measured GRF magnitudes from the kinetic dataset.}\label{tab:grf}%
\begin{tabular}{@{}lcc@{}}
\toprule
Method & Mean Error (\unit{\newton}) & \unit{\percent} of Peak\\
\midrule
sinusoidal GRF & $171.3$ & $17.16$\\
freeform GRF & $471.3$ & $47.23$\\
\bottomrule
\end{tabular}
\end{center}
\end{table}

The sinusoidal \kfte makes a drastic improvement on the GRF error that is evident in both Table~\ref{tab:grf} and Figure~\ref{fig:grf_estimation}.
From the example in the figure, the sinusoidal \kfte provides good tracking of the GRF trajectory, whereas the freeform \kfte fails to produce a valid GRF trajectory.
\section*{Discussion}
The kinetic dataset facilitated the development of a whole-body dynamic state estimator using the \kfte method.
From Table~\ref{tab:grf}, it is clear that the sinusoidal \kfte produces more accurate GRF estimates.
The GRF estimate resembles that of the force measurements from the force plates (see Figure~\ref{fig:grf_estimation}(\subref{fig:grf_1})).
Many of the state-of-the-art physics-based methods have validated GRF estimate errors in a similar fashion, by using existing studies on predetermined locomotion styles~\cite{shimada2020physcap, rempe2020contact}.
However, one study validated its GRF estimates using a parkour dataset~\cite{li2022estimating}, which quotes similar mean errors to our work (see Table~\ref{tab:grf}).

Furthermore, this result confirms the use of the assumed sinusoidal GRF profile for steady-state galloping of quadrupeds.
This was previously applied to horses, where it was also found to be a good fit for the data~\cite{witte2004determination}.
Limiting the GRF to a sinusoidal profile allowed for an easier optimisation problem to solve, and reduced the problem of non-unique solutions by restricting the space of possible GRFs to those matching a widely-observed form from previous quadruped studies.


Note that although the GRF error is vastly different between the two methods reported in Table~\ref{tab:grf}, the resultant 3D reconstructions are similar (calculated average MPE of \qty{13.0}).
Therefore, the freeform \kfte does yield forces with plausible magnitudes even if the profile does not match the observed truth data.
We compared the ground truth impulse to the estimated impulse for both methods for the same trial (T4 of the kinetic dataset).
The ground truth impulse was calculated as \qty{119.4}{\kilo\gram\metre\per\second}, and the sinusoidal \kfte produced an impulse of \qty{122.32}{\kilo\gram\metre\per\second}, whereas the an impulse of \qty{97.50}{\kilo\gram\metre\per\second} was obtained for the freeform \kfte.
Furthermore, this is confirmed by the GRF trajectory in Figure~\ref{fig:grf_estimation}(\subref{fig:grf_2}), where the area under the curve (\ie the impulse) is similar to the ground truth ($< \qty{20}{\percent}$ error).

The estimated joint torques produced a consistent torque profile on both datasets (see Figures~\ref{fig:kinetic_torque_analysis} and~\ref{fig:acinoset_torque_analysis}).
This result is in line with the results on the hindlimbs of racing greyhounds (similar to the cheetah in that they are very fast quadrupeds)~\cite{williams2009exploring}.
However, the quoted torques are of different magnitudes, with the greyhound's back ankle average peak roughly calculated as \qty{40}{\newton\metre}, while the cheetah's average peak was calculated to be around \qty{100}{\newton\metre}.
This is roughly twice the torque generated during the stance when compared to the greyhound.
It is known that the cheetah generates larger joint torques to resist larger GRFs than the greyhound~\cite{hudson2011functional}.
However, it would not be advisable to place too much emphasis on this particular result given the simplified kinematic model used (in addition to inverse dynamics) to generate these quantities in this work.

The 3D pose estimation is adequate and has not drastically changed from the previous \acinoset baseline~\cite{muramatsu2022improving}, as shown in Table~\ref{tab:torques} and Figure~\ref{fig:qualitative}.
The MPE is very low and suggests that adding a more complete physics model has not diminished the resultant kinematics.
Therefore, without compromising our pose estimation accuracy, we have gained kinetic information, and with it, a more complete picture of the dynamics of the cheetah.

\begin{figure}[htpb]
\centering
\includegraphics[scale=0.25]{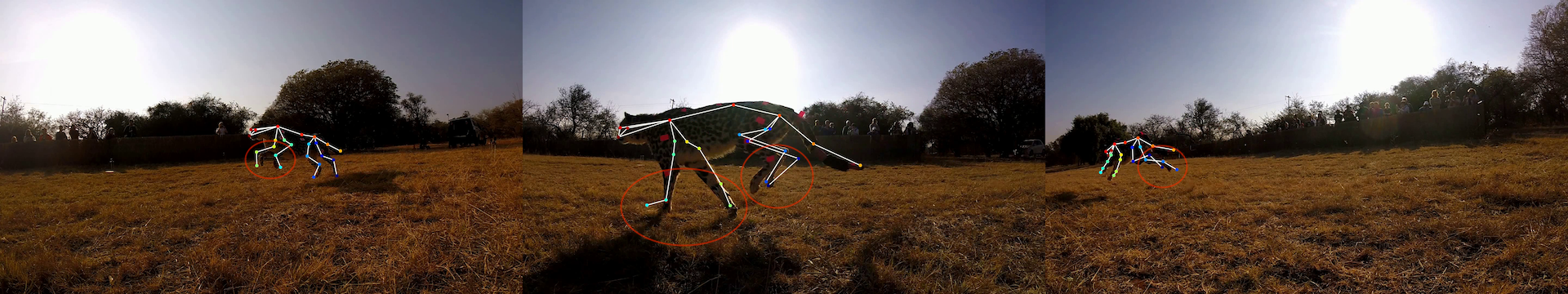}
\caption{Visual examples of bad skeleton fits when projected onto the image plane after dynamic estimation. The red circles show clear pose estimation issues, the main contributor being the rigid body model failing to adapt to the extension/contraction of the forelimbs during locomotion.}\label{fig:qual-bad}
\end{figure}

That said, an average PCK of \qty{62.94}{\percent} suggests that there is room for improvement in our 2D pose estimation.
This result could be attributed to the shortcomings of using a simplified rigid body model for the cheetah.
The rigid body model, together with fixed joint centres, are not representative of the flexibility within the cheetah's skeleton.
In addition, modelling the muscles as torque-driven actuators is clearly not characteristic of a cheetah's muscle system.
Lastly, point contacts are used to model the dynamics between the foot and the ground, which completely overlooks the complexity of the cheetah's foot/paw. As highlighted in Figure~\ref{fig:qual-bad}, there are times when a rigid body model cannot accommodate the forelimbs and neck body parts because of a lack of extension/contraction.

Nevertheless, the developed method provides a stepping stone to accurate dynamic analysis of wild quadrupeds using a non-invasive multi-camera system. Ultimately, the accuracy of the estimated 3D kinematics were adequate for our purposes.

Looking to the future, we have identified two aspects of the current implementation that we could improve: removing the explicit assumption on the GRF profile, and the rigid body model. First, we would like to investigate the incorporation of a learned prior model of the GRF profile in the cost function as a possible improvement to the existing freeform GRF method.  The probability model can use examples from existing force data on quadrupeds.  This would enable the study of more dynamic movements (turns and rapid acceleration or deceleration). 

Second, even though the overall 3D kinematic estimate was sufficient for this work, we can potentially obtain more accurate predictions by remodelling the shoulder and neck links as variable length~\cite{fukuhara2020bio} which should address the problem highlighted in Figure~\ref{fig:qualitative}.
 
\section*{Methods}
We first provide background theory to multi-body dynamics and the \kfte method used in this work.
Then, we provide more details on the the kinetic and \acinoset datasets.
Lastly, we present the evaluation metrics used in this work.
\subsection*{Multi-body Dynamics}
The dynamics of the cheetah was modelled as a rigid multi-body system using absolute angle coordinates $\mathbf{q}(t)$~\cite{knemeyer2020minor}, often expressed as
\begin{equation}
    \mathbf{M}(\mathbf{q})\mathbf{\ddot{q}} + \mathbf{C}(\mathbf{q}, \mathbf{\dot{q}})\mathbf{\dot{q}} = \boldsymbol{\tau}_{g}(\mathbf{q}) + \mathbf{B}\mathbf{u} + \mathbf{J}^{T}_{L}(\mathbf{q})\boldsymbol{\lambda} + \mathbf{J}^{T}_{c}(\mathbf{q})\mathbf{T}_{c}\label{eqn:dynamics},
\end{equation}
where $\mathbf{M}(\mathbf{q})$ represents the inertia matrix, $\mathbf{C}(\mathbf{q}, \mathbf{\dot{q}})$ captures Coriolis and centrifugal terms, $\boldsymbol{\tau}_{g}(q)$ is the gravity vector, $\mathbf{B}$ maps $\mathbf{u}$ inputs to generalised forces, $\mathbf{J}^{T}_{L}(\mathbf{q})$ is the contact Jacobian and $\boldsymbol{\lambda}$ the corresponding contact forces, and $\mathbf{J}^{T}_{c}(\mathbf{q})$ is the angle constraint Jacobian that maps the constraint torques $T_{c}$ into the equation of motion.

In this work, $\mathbf{B}\mathbf{u}$ represents the joint torques produced by the cheetah during locomotion, henceforth denoted by $\boldsymbol{\tau}$.

The rigid multi-body model of the cheetah consists of \qty{17} interconnected links, resulting in a state vector $\mathbf{q}$ of size \qty{54} (\qty{3} angles for each link and an additional \qty{3} for the position of the base link in the inertia frame).

Each link was modelled as a cylinder described by three parameters, length, mass, and radius. These parameters were determined per subject, using the dimensions and masses quoted in multiple resources produced Hudson et al.~\cite{hudson2012high, hudson2011functional, hudson2011functional2}.
Note that the length parameters were fine-tuned by projecting the cheetah skeleton into one of the cameras and validating that the links match the limb lengths of the cheetah.
\subsection*{Trajectory Optimisation}
In general, the trajectory optimisation problem can be described as a non-linear program (NLP). Here, we formulate a NLP with the following constraints:
\begin{equation}
    \underset{\text{$\mathbf{q}$, $\boldsymbol{\tau}$,$\boldsymbol{\lambda}$,$\boldsymbol{T_{c}}$}}{\text{min}} \quad \sum^{N}_{k=0}g(\mathbf{q}_{k},\boldsymbol{\tau}_{k})
    \label{eqn:traj_min}
\end{equation}
subject to
\begin{align}
   \mathbf{\dot{q}} = f(\mathbf{q},\boldsymbol{\tau},\boldsymbol{\lambda}), \label{eqn:dynamics_constraints} \\
    C(\mathbf{q},\boldsymbol{\lambda}) = 0, \label{eqn:contact_constraints}
\end{align}
where $f(\cdot)$ denotes the multi-body dynamics (Equation~\eqref{eqn:dynamics}) and $C(\cdot)$ are the path and contact constraints. The cost function to optimise is $g(\cdot)$. 
Note that constraints (\ref{eqn:dynamics_constraints}) and (\ref{eqn:contact_constraints}) must be satisfied for the duration of the trajectory $k \in [0, N]$.

First, an equality constraint is used to enforce the dynamics of the system (\ie $f(\vect{q}_{k}, \boldsymbol{\tau}_{k}, \boldsymbol{\lambda}_{k}, \vect{a}_{k})$, see Equation~\eqref{eqn:dynamics}):
\begin{equation}
    \vect{M}_{k}\ddvect{q}_{k} + \vect{C}_{k}\dvect{q}_{k} - (\vect{G}_{k} + \boldsymbol{\tau}_{k} + \vect{J}^{T}_{L,k}\boldsymbol{\lambda}_{k} + \vect{J}^{T}_{A,k}\vect{a}_{k}) - \vect{w}_{k} = \vect{0},\label{eq:traj_dynamic_2}
\end{equation}
defined at each finite time $k$.
The function notation has been omitted for clarity (\eg $\vect{M}(\vect{q})$ is simply stated as $\vect{M}$).

In addition to the motion constraint above, two constraints to incorporate the measurement data from the cameras are specified:
\begin{enumerate}
    \item A constraint that relates the current pose $\vect{q}$ to a set of 3D marker positions.\label{en:3d_markers}
    \item A constraint that relates reprojection of the 3D marker positions to the 2D pixel coordinates for each camera.\label{en:2d_pixels}
\end{enumerate}
Constraint~\ref{en:3d_markers} can be determined using the pose equations that relate the generalised coordinates to positions.
This is defined as a general function $g(\vect{q}_{k})=\vect{x}_{k}$ made into an equality constraint, $g(\vect{q}_{k}) - \vect{x}_{k} = \vect{0}$, at finite time $k$.
Constraint~\ref{en:2d_pixels} uses a camera projection matrix, $\vect{P}$, to transform a 3D marker from world coordinates to 2D image space.
The aim is to minimise the error between the ground truth and reprojected pixel locations; therefore, $\vect{v}$ is used to capture this error so that it can be added to the cost function.
The resultant equality constraint is defined as
\begin{equation}
    \vect{y}_{k,c,m} - \vect{P}_{c}\vect{x}_{k,m} - \vect{v}_{k,c,m} = \vect{0},
\end{equation}
where $c$ selects a specific camera to project to, $m$ selects a specific marker on the cheetah, and $k$ is the finite time that this operation is performed at.

The constraints used in the optimisation include not only the measurement data but also two additional types of constraints: contact constraints to enforce the correct GRF, represented by $\boldsymbol{\lambda}_{k}$, and joint constraints to ensure that constraint torques, $\vect{a}_{k}$, to prevent motion in forbidden directions (which is necessary when using absolute angle coordinates).

A no-slip model of contact is developed using the foot velocity $\vect{v}_{f}$ and contact force $\boldsymbol{\lambda}$ to construct the following constraints,
\begin{align}
    \boldsymbol{\lambda} > 0,\\
    |\vect{v}_{f}| \leq \epsilon,\label{eq:contact_mag}
\end{align}
where the first constraint enables the generation of a GRF while the second constraint ensures the foot is approximately fixed ($\epsilon \leq 1$) to the ground during the contact phase.
The contact constraints are relaxed to within a slack error, $\epsilon$, to ensure that the optimisation problem is easier to solve.

Since sliding contact is not considered, the contact force $\boldsymbol{\lambda}$ has to satisfy friction cone constraints~\cite{trinkle1997dynamic} to estimate valid forces.
To obtain a simpler model of contact, linearised friction cone constraints are formulated (resulting in a friction pyramid):
\begin{align}
    |\lambda_{x}| &\leq \mu_{s}\lambda_{z},\\
    |\lambda_{y}| &\leq \mu_{s}\lambda_{z},
\end{align}
where $\lambda_{x}$ and $\lambda_{y}$ are the tangential components of contact force, $\lambda_{z}$ is the vertical component ($\lambda_{z} > 0$), and $\mu_{s}$ is the friction coefficient (and was set to \num{1.3} for all experiments~\cite{wilson2013locomotion}).
The tangential components of the contact force are decomposed into two positive variables in implementation, \ie $\lambda_{x}=\lambda_{x}^{+} - \lambda_{x}^{-}$ where $\lambda_{x}^{+},\lambda_{x}^{-}\geq0$.
This facilitates the optimisation by replacing the non-linear absolute value function with a linear version (\eg $|\lambda_{x}|\equiv \lambda_{x}^{+} + \lambda_{x}^{-}$)\footnote{This is only true if either $\lambda_{x}^{+}$ or $\lambda_{x}^{-}$ is zero. IPOPT ensured that this condition was true.} in the above friction cone constraints.

There is a requirement for explicit modelling of joints to ensure the correct usage of absolute angle coordinates.
Two different types of joints are supported: revolute and universal.
The former ensures that the constraints
\begin{align}
    \vect{r}_{y}\cdot\vect{r}_{x} &= 0,\\
    \vect{r}_{y}\cdot\vect{r}_{z} &= 0,
\end{align}
are met for when the joint is restricted to rotate about the y-axis only, removing two degrees of freedom.
Note that $\vect{r}_{y}$ is the column vector from the rotation matrix $\vect{R}$ that associates rotations about the y-axis.
The latter ensures that the following constraint is met
\begin{equation}
    \vect{r}_{y}\cdot\vect{r}_{x} = 0,
\end{equation}
preventing a rotation about the z-axis and removing a single degree of freedom.

Finally, the cost function to be minimised is defined as
\begin{equation}
    g(\vect{q}, \boldsymbol{\tau}) = \alpha_{1}e_{\text{\it meas}} + \alpha_{2}e_{\text{\it model}} + \alpha_{3}e_{\text{\it smooth}},\label{eqn:traj_cost_2}
\end{equation}
where $\alpha_{1}=1$, $\alpha_{2}=\num{10000}$, and $\alpha_{3}=1$. The $e_{\text{\it meas}}$ term is defined as
\begin{equation}
    e_{\text{\it meas}} = \sum_{k=1}^N \sum_{j=1}^c \sum_{i=1}^m \sum_{l=1}^2 C\bigg(\frac{\mathbf{v}_{k,j,i,l}}{\sigma_{i}}\bigg),
    \label{eqn:traj_cost_meas}
\end{equation}
where $C(\cdot)$ is the redescending robust cost function~\cite{costArticle} and $\sigma_{i}$ is used to normalise the measurement error.
This uncertainty parameter is derived from the predicted 2D keypoint error distribution.
The uncertainty parameters were obtained in a similar fashion to our previous work~\cite{joska2021acinoset}.
In addition, $e_{\text{\it model}}$ now represents the error in the dynamic equation of motion instead of the assumed constant acceleration motion:
\begin{equation}
    e_{\text{\it model}} = \sum_{i=1}^n \sum_{j=1}^p\mathbf{w}_{i,j}^2.
    \label{eqn:traj_cost_model_2}
\end{equation}
Note that the above error did not require normalisation.

Lastly, the $e_{\text{\it smooth}}$ term is defined as:
\begin{equation}
    e_{\text{\it smooth}} = \sum_{k=1}^n \sum_{l=1}^t 10(\boldsymbol{\tau}_{k,l})^2 + 0.1h^2\sum_{k=1}^n \sum_{i=1}^m (\ddvect{x}_{k,i})^2,\label{eqn:smoothness_term}
\end{equation}
where $h$ is the frame rate.
The first term penalises the joint torques to ensure a minimum energy solution.
The literature suggests that it is a common assumption that humans and animals conserve energy during movement~\cite{brubaker2009physics}.
The \qty{10} weight is used to increase its contribution during minimisation. 

During experiments, it was noted that the optimal solutions would produce jittery motion.
For this reason, the second term in Equation~\eqref{eqn:smoothness_term} was added to the cost function.
This term penalises the acceleration of the 3D markers, favouring solutions that have smooth trajectories for all 3D markers.
The $h^2$ weighting term performs normalisation, and the \num{0.1} weight is used to decrease its contribution during minimisation.
Note that through normalisation, each term is assumed to be equally weighted in Equation~\eqref{eqn:traj_cost_2}.
However, the goal is for the resultant motion estimate to be as physically plausible as possible, and therefore an enormous weight is placed on the $e_{\text{\it model}}$ to direct the optimiser to find solutions that essentially force the model noise to be zero, \ie $\vect{w}=\vect{0}$.
Note that IPOPT~\cite{IPOPTarticle} was used and configured with the MA97 linear solver~\cite{hsl2007collection} to implement the optimisation.
\subsubsection*{\kfte}
\begin{figure}[htbp]
\centering
\includegraphics[scale=0.6]{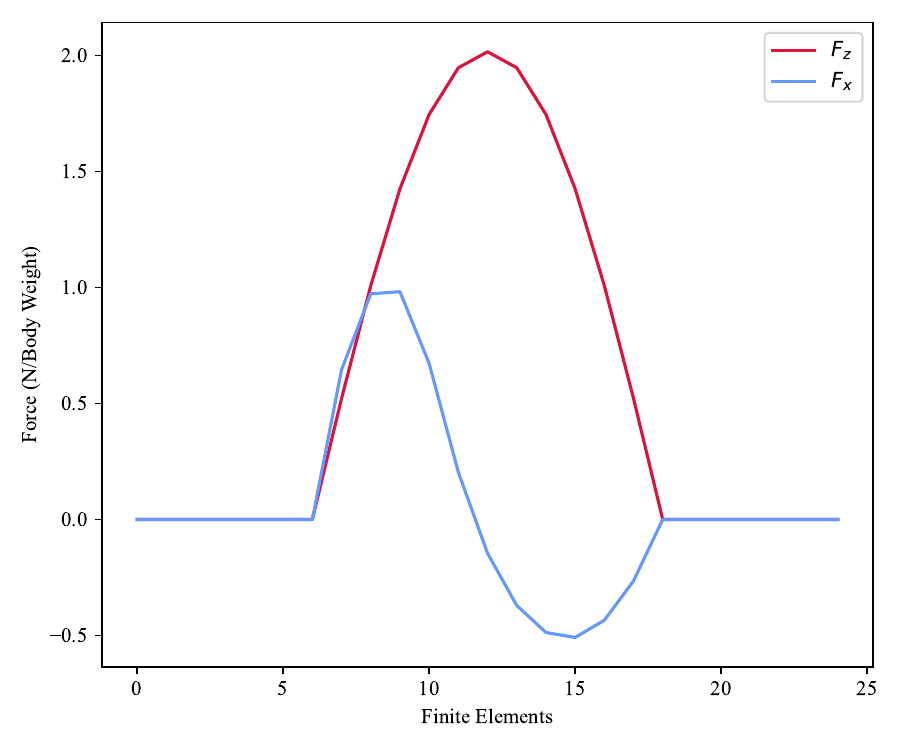}
\caption{An example of the synthesis of the sinusoidal GRF profile.}\label{fig:sin_grf}
\end{figure}

Trajectory optimsiation was used to investigate two \kfte methods: sinusoidal GRF and freeform GRF.
The sinusoidal \kfte assumes a ``sinusoidal'' GRF profile that essentially reduces the optimisation problem to solely estimate the joint torques (we allow for \qty{20}{\percent} variation from this assumption).
While the freeform \kfte does not make any assumptions about the GRF profile, it instead performs a complete joint estimation of the torques and GRFs.
Both methods set up the same optimisation problem (as outlined above)\footnote{We assume contact timing is known and therefore did not require an automatic contact detection algorithm.}, but the sinusoidal \kfte has an extra pre-processing step that generates the initial GRF profile.
An example of the expected profile is shown in Figure~\ref{fig:sin_grf}.

In Figure~\ref{fig:sin_grf}, the $F_{z}$ component is created using a half sine wave with an amplitude that is determined through a linear model for each limb.
The linear model relates the running speed of the cheetah with the peak vertical force~\cite{hudson2012high}.
The $F_{x}$ component is approximated by a spline using 5 control points: zero points for the start, midpoint, and end of the stance, $\frac{F_{z}}{2}$ for the deceleration peak, and $\frac{F_{z}}{4}$ for the acceleration peak.
We assume that there is no $F_{y}$ component present in the straight-line steady-state gallops that are included in the datasets (in every trial, the cheetah runs along the x-axis).

\subsection*{Datasets}
Two cheetah locomotion datasets were used to develop and evaluate the proposed approach. Neither was a perfect test case -- one (AcinoSet) lacked force plate data, while the other used a non-standard camera calibration that resulted in slightly higher uncertainty in the animal's pose -- however, the combination of the two with these complimentary deficiencies was sufficient to prove the concept. The availability of more complete and accurate locomotion data will likely improve this method in future. 

\subsubsection*{AcinoSet}
\acinoset~\cite{joska2021acinoset} is a cheetah-running dataset that is used for the evaluation of the motion and torque estimation.
The dataset contains 90 running videos with six different views from low-cost GoPro cameras (\numproduct{2704 x 1520} resolution filmed at \num{120} frames per second (FPS) and \numproduct{1920 x 1080} resolution, filmed at \qty{90}{FPS}).
There are \num{7588} human-annotated frames and the average video length is approximately \qty{2}{\second}.
Furthermore, the dataset includes \dlc~\cite{mathis2018deeplabcut} 2D pose estimates for each camera view for each video set, as well as the corresponding 3D reconstruction data, in addition to both the intrinsic and extrinsic calibration data.
The intrinsic camera calibration data included fisheye lens distortion parameters.

Here, we use the \dlc network that was developed in our previous work~\cite{muramatsu2022improving} and a subset of five trials from \acinoset for evaluation (see Table~\ref{tab:acinoset_dataset}).
Each trial consists of the cheetah doing a steady-state gallop for two different subjects.
\begin{table}[htbp]
\begin{center}
\caption{The test dataset selected from the \acinoset. The length is presented as the number of frames in the trajectory divided by the video frame rate. Each trajectory length is equivalent to one stride performed by the cheetah.}\label{tab:acinoset_dataset}
\begin{tabular}{lll}
\toprule
Test & Trial & Length (\si{\second})\\
\midrule
T1 & 2017\_08\_29/top/phantom/run1\_1 & $44/90$ \\
T2 & 2017\_08\_29/top/jules/run1\_1 & $30/90$ \\
T3 & 2017\_09\_02/top/jules/run1 & $30/90$ \\
T4 & 2019\_03\_07/phantom/run & $57/120$ \\
T5 & 2017\_09\_02/bottom/jules/run2 & $33/90$ \\
\bottomrule
\end{tabular}
\end{center}
\end{table}
\subsubsection*{Kinetic Dataset}
\begin{figure*}[htbp]
\centering
\begin{subfigure}{0.5\textwidth}
  \centering
  \includegraphics[width=\textwidth]{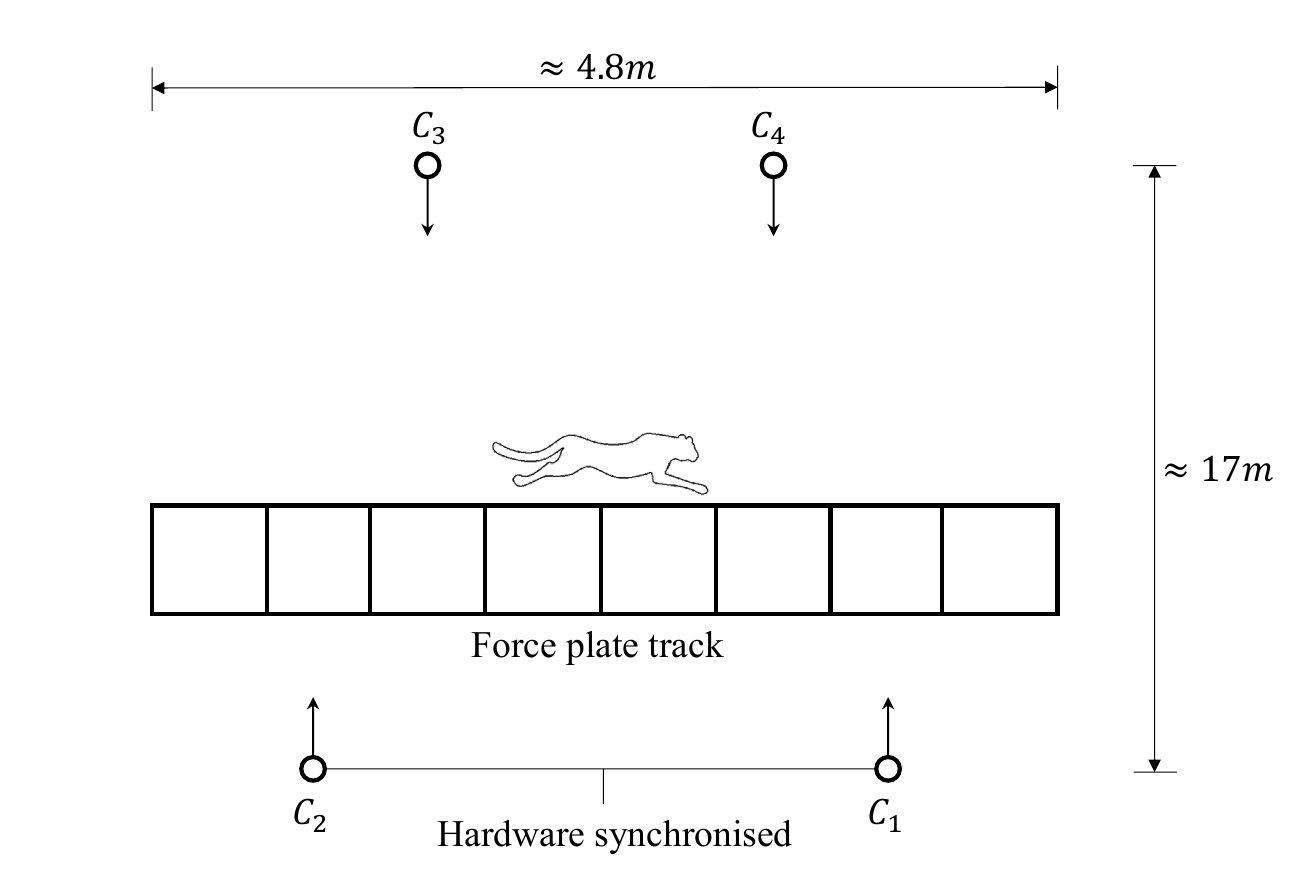}
  \caption{}
  \label{fig:kinetic_scene}
\end{subfigure}%
\begin{subfigure}{0.5\textwidth}
    \centering
    \includegraphics[width=\textwidth]{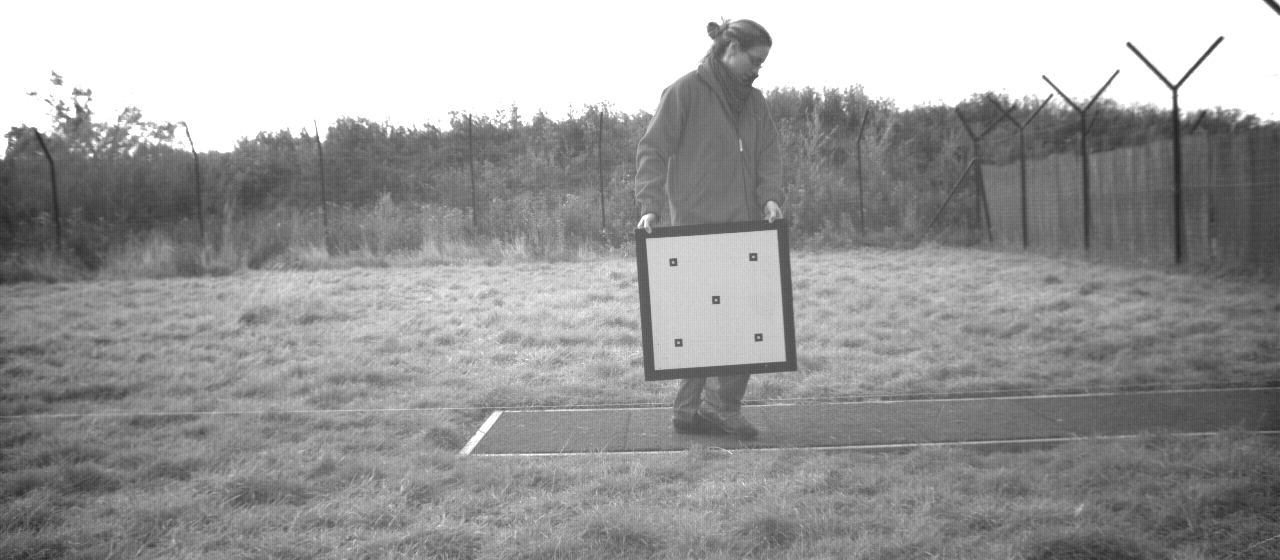}
    \caption{}
    \label{fig:kinetic_calib}
\end{subfigure}
\begin{subfigure}{0.5\textwidth}
  \centering
  \includegraphics[width=\textwidth]{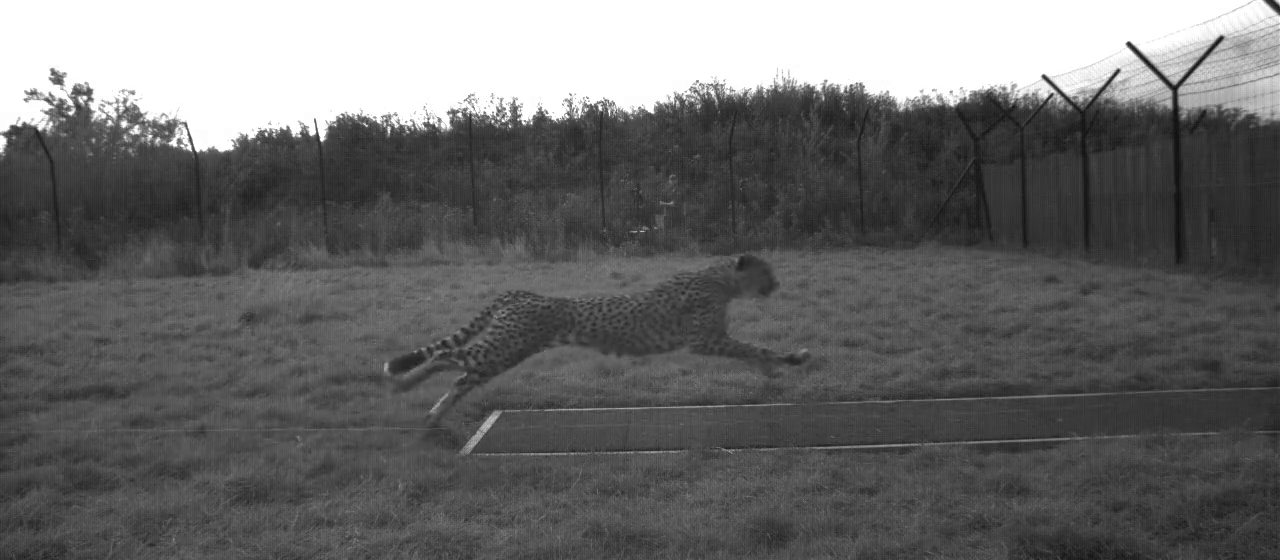}
  \caption{}
  \label{fig:example_kinetic_frame}
\end{subfigure}%
\begin{subfigure}{0.5\textwidth}
    \centering
    \includegraphics[width=\textwidth]{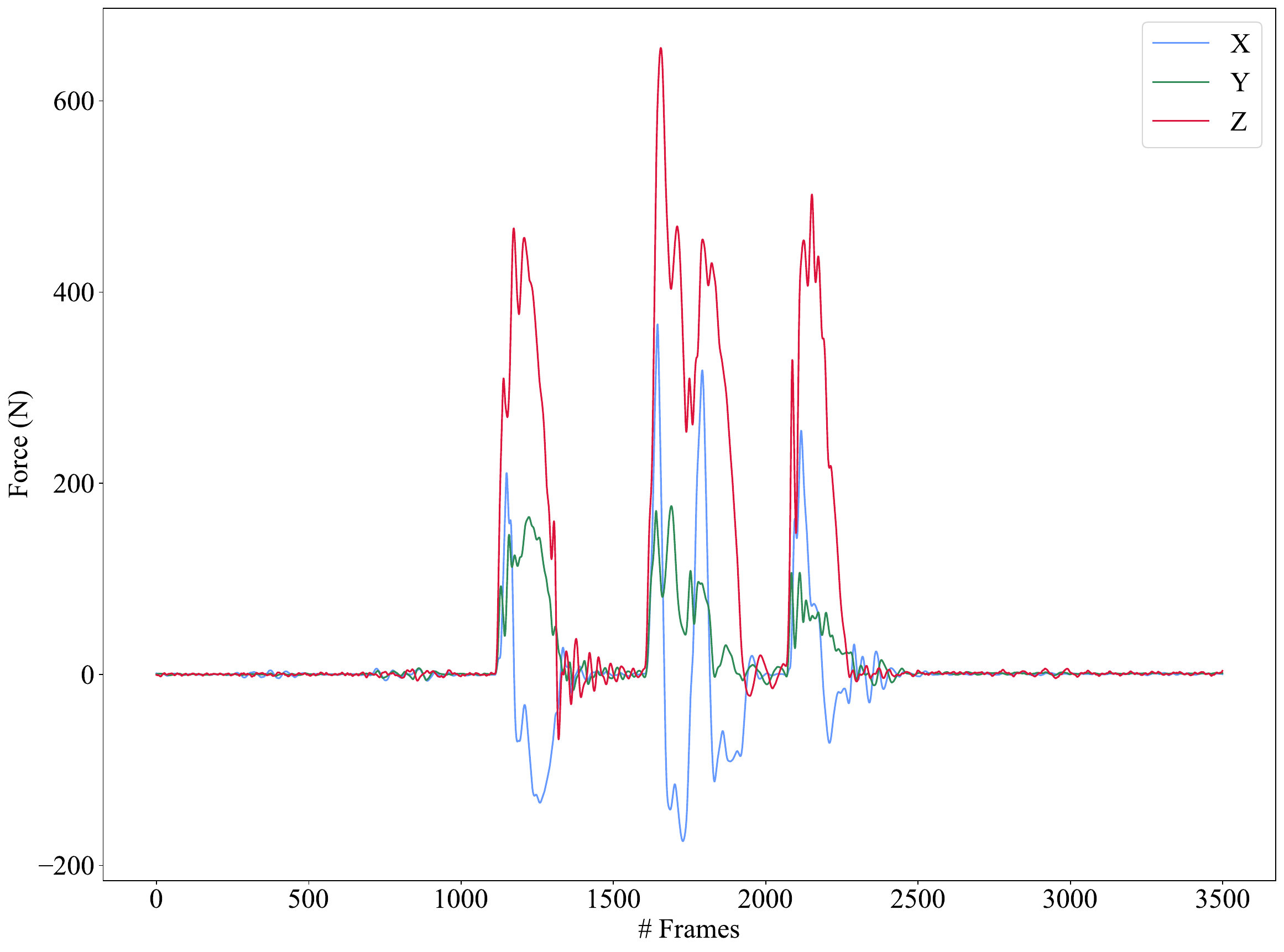}
    \caption{}
    \label{fig:kinetic_grf}
\end{subfigure}
\caption{Example data from the kinetic dataset: (\subref{fig:kinetic_scene}) is a diagram of the scene setup, (\subref{fig:kinetic_calib}) provides an example of the calibration target used to perform intrinsic and extrinsic calibration, (\subref{fig:example_kinetic_frame}) shows a frame from the captured video, and (\subref{fig:kinetic_grf}) shows the measurement data from the force plates.}
\label{fig:kinetic_setup}
\end{figure*}

This dataset was generated for this work to validate the proposed method of dynamic data estimation through the use of force plate measurements.
The dataset was acquired from The Royal Veterinary College in their previous work~\cite{hudson2012high}.
The dataset contains grayscale videos (including calibration targets) filmed at \qty{1000}{FPS} (\numproduct{1280 x 560} and \numproduct{800 x 600} resolutions) with synchronised force plate data sampled at \qty{3500}{\hertz} (see Figure~\ref{fig:kinetic_setup}).
From the original dataset, a subset of five trials of two different subjects\footnote{The subjects were \textit{Cheetah 1} and \textit{Cheetah 2} referred to in Table 1 of the paper by Hudson et al. ~\cite{hudson2012high}.} was selected, and is denoted the `'kinetic dataset'' in this work (see Table~\ref{tab:kinetic_dataset}).
The original work gathered trials of the cheetahs chasing a mechanical lure across eight Kistler force plates (model 9287A and 9287B) and in front of four cameras~\cite{hudson2012high}.
Both the video and the force plate data were resampled to a \qty{200}{\hertz} sample rate for the kinetic dataset.
This allows for a more tractable \kfte problem formulation.

As the dataset only consisted of grayscale videos and synchronised force plate measurements, 2D keypoint inference and a scene calibration are required to enable 3D reconstructions of the cheetah.
\begin{table}[htbp]
\begin{center}
\caption{The test dataset selected from the kinetic dataset. The length is presented as the number of frames in the trajectory divided by the video frame rate. Each trajectory length is equivalent to one stride performed by the cheetah.}\label{tab:kinetic_dataset}
\begin{tabular}{lll}
\toprule
Test & Trial & Length (\si{\second})\\
\midrule
T1 & 2009\_09\_07/arabia/trial06 & $49/200$\\
T2 & 2009\_09\_07/shiraz/trial04 & $52/200$\\
T3 & 2009\_09\_08/shiraz/trial04 & $52/200$\\
T4 & 2009\_09\_11/shiraz/trial01 & $55/200$\\
T5 & 2009\_09\_11/shiraz/trial02 & $48/200$\\
\bottomrule
\end{tabular}
\end{center}
\end{table}
\subsubsection*{2D Pose Estimation}
The \dlc~\cite{mathis2018deeplabcut} network trained for \acinoset (RMSE of \qty{8.38}{pixels} on the test set) was first used to obtain 2D keypoints on the cheetah in this dataset.
However, to quantify the networks ability to infer 2D keypoints on an entirely new dataset, we needed to produce a new test set from the kinetic dataset.
A subset of \num{260} frames of the new data was hand-labelled to form a test set, and the RMSE between ground truth and inferred 2D keypoints was \qty{10.85}{pixels}.
In order to reduce the error on the test set, we decided to train a new network to improve inference on the kinetic dataset.
An additional \num{95} frames were manually labelled from the kinetic dataset to improve the training dataset for \acinoset.
This improved performance on the kinetic test set by reducing the RMSE to \qty{6.10}{pixels}.
\subsubsection*{Camera Calibration}
The calibration techniques developed for \acinoset were used to perform intrinsic and extrinsic calibration for the kinetic dataset.
There were some hurdles to overcome, in that the dataset did not contain a standard checkerboard calibration target that was compatible with OpenCV's calibration library~\cite{opencv_library} (see Figure~\ref{fig:kinetic_setup}(\subref{fig:kinetic_calib})).
This meant that the automatic corner finder could not be used.
Instead, multiple frames of the calibration target from each camera were gathered and the points of interest were hand-labelled.
There were \num{9} points of interest that were chosen on the calibration target shown in Figure~\ref{fig:kinetic_setup}(\subref{fig:kinetic_calib}): the \num{4} inside corners of the black border, \num{4} centre points of the outer squares, and the single centre point of the centre square.
The hand labelling of the calibration target was done for roughly \num{16} frames across the camera's field of view, and for each trial included in the dataset.
Once the hand labelling was done, a standard intrinsic calibration was performed to obtain the camera intrinsic parameters.
Note that radial distortion parameters were estimated during the intrinsic camera calibration process.

The extrinsic calibration of all four cameras simultaneously proved difficult due to the lack of shared scenes between the cameras and the calibration target.
In particular, the kinetic dataset used in the calibration process contained videos of the calibration target that were not seen in more than two cameras at the same time, and the calibration target was not viewed at the same time across the force plate track.
This meant that there was no shared scene data between the first stereo pair ($C_{1}$ and $C_{2}$) and the second stereo pair ($C_{3}$ and $C_{4}$) shown in Figure~\ref{fig:kinetic_setup}(\subref{fig:kinetic_scene}).
Based on the distance from the force plate track, cameras $C_{1}$ and $C_{2}$ are referred to as the near side stereo pair, and cameras $C_{3}$ and $C_{4}$ are referred to as the far side stereo pair.

To overcome this challenge, a stereo calibration was performed between cameras $C_{1}$ and $C_{2}$ to obtain the pose of camera $C_{2}$ relative to camara $C_{1}$.
The world frame was set to the first force plate, and the pose of camera $C_{1}$ was determined using the perspective-n-point (PnP) algorithm~\cite{opencv_library}.
This algorithm uses a set of 3D-2D point correspondences on the force plates, viewed by camera $C_{1}$, to compute the pose of the camera relative to the world frame.

After this stereo calibration, the extrinsic parameters for the near side stereo pair had been estimated.
However, it was not possible to determine the pose of the far side cameras relative to the near side due to the lack of shared scenes of the calibration target.
To overcome this limitation, a joint optimisation process was utilised to obtain the pose of the cameras on the far side.
By using the points on the cheetah, and assuming the cheetah runs in a straight line, it was possible to set up an optimisation problem that jointly estimated the cheetah's 3D pose and the pose of one of the far side cameras.
The stereo calibration of the near side cameras was enough to seed the optimisation for roughly \num{30} frames of an example trial.
The optimisation cost function was defined to minimise reprojection error.
This process was performed for each camera on the far side separately.
The optimal solutions from the optimisation were found to be reasonable by comparing them with the rough measurements that were taken of the scene by the original researchers~\cite{hudson2012high}.

It is worth noting that this non-standard approach added uncertainty to the observations made by the far side cameras.
Therefore, when using \kfte for 3D reconstructions of the kinetic dataset, an extra \qty{40}{\percent} (\ie multiply observation errors by \num{0.6}) of uncertainty was added to the observation errors from the far side cameras.
This figure was found to provide good solutions during experimentation.

\subsection*{Evaluation}
\subsubsection*{Root-mean-square Error}
The RMSE is a general error metric used to quantify the performance of a model by measuring the differences between estimated $\vect{\hat{x}}$ and ground truth $\vect{x}$ values in $\mathbb{R}^{n}$. It is defined as
\begin{equation}
    e_{\!_{\text{RMSE}}} = \sqrt{\frac{1}{n}\sum_{i=1}^{n}(x_{i} - \hat{x}_{i})^{2}},
\end{equation}
where $x_{i}$ is the i\textsuperscript{th} element of the vector $\vect{x}$.
\subsubsection*{2D Metrics}
To assess pose estimation methods in image space, two reprojection error metrics are used: $\ell^2$ norm in pixels and percentage correct keypoints (PCK)~\cite{andriluka20142d}.
The $\ell^2$ norm of a vector $\vect{x}\in\mathbb{R}^{n}$ is defined as
\begin{equation}
    \lVert\vect{x}\rVert_{2} = \sqrt{\sum_{i=1}^{n}|x_{i}|^{2}}.
\end{equation}
PCK is a metric that considers the percentage of estimated keypoints that are correctly located within a certain threshold distance of the ground truth keypoints.
In this project, a matching threshold of the nose-to-eye(s) segment length was used, similar to what was done in a study by Mathis et al. ~\cite{mathis2021pretraining}.
This value is a normalised error metric that is calculated in pixels from the ground truth keypoints.
Note that the PCK metric is `strict', in that the segment length is small relative to the body and therefore considers a small region for a correct keypoint.
In addition, this metric is only evaluated on images where the nose and one of the eyes are clearly visible.
This allows the threshold to be determined.

Unless stated otherwise, the $\ell^2$ norm is used to denote a `general error' value for prediction performance.
\subsection*{3D Metrics}
To assess 3D pose estimation methods, a 3D position error metric is used: global mean position error (MPE) over the full trajectory.
The MPE accounts for the absolute position in 3D space and is defined as 
\begin{equation}
    e_{\!_{\text{MPE}}} = \frac{1}{NM}\sum_{k=1}^{N}\sum_{j=1}^{M}\lVert\vect{x}_{k,j}-\vect{\hat{x}}_{k,j}\rVert,\label{eq:mpe}
\end{equation}
where $N$ is the length (in frames) of the trajectory, $M$ is the number of markers (placed on the joints of the cheetah), $\vect{x}_{k,j}$ is the position of marker $j$ in $\mathbb{R}^{3}$ at time $k$, and $\vect{\hat{x}}_{k,j}$ is an estimate of $\vect{x}_{k,j}$.

\section*{Conclusions}
In this paper, we present kinetic full trajectory estimation: a model-based nonlinear optimisation approach that incorporates estimation of the ground reaction forces to improve markerless 3D pose estimation in wild animals. Although this cannot be regarded as a remote force measurement method, the GRF profiles obtained using the sinusoidal model agree with those observed in this study from force plate data obtained from running cheetahs, as well as other studies of running quadrupeds. The approach produced promising results when tested on running cheetahs. In future, we hope that access to more complete truth data, and the incorporation of more advanced ground reaction force and skeletal models will improve the approach sufficiently to support its use in the study of fast, dynamic manoeuvres. 

\bibliography{main}

\section*{Data availability}
All data and code are available at \href{https://github.com/African-Robotics-Unit/torque-estimation}{https://github.com/African-Robotics-Unit/torque-estimation}.

\section*{Author contributions statement}
Z.D.S.\ performed algorithm design, experiments and primary writing of the manuscript. S.S.\ assisted in algorithm design and review/revision of the manuscript. A.P. conceived the study and revised the manuscript. P.E.H. and A.M.W. collected the cheetah kinetic data, and F.N.\ provided technical inputs on the image processing and reviewed the manuscript.

\end{document}